\documentclass[runningheads]{llncs}
\usepackage{fmtcount}
\usepackage[colorlinks,linkcolor=blue]{hyperref}
\usepackage{graphicx,verbatim}
\usepackage{amssymb}
\usepackage{amsmath}
\usepackage{amssymb,bbding}
\usepackage{multirow}
\usepackage{tabularx}
\usepackage{url}
\usepackage[table,xcdraw]{xcolor}
\usepackage{amsmath}
\usepackage{xcolor}
\usepackage{graphicx} 
\usepackage{booktabs}
\usepackage{mathrsfs}

\usepackage{marvosym}
\usepackage{color}

\usepackage[marginal]{footmisc}

\begin{document}
\title{Medical-Knowledge Driven Multiple Instance Learning for Classifying Severe Abdominal Anomalies on Prenatal Ultrasound}
\titlerunning{MIL for Classifying Abdominal Anomalies on Prenatal US}
%

\author{Huanwen Liang\inst{1} \and Jingxian Xu\inst{1} \and Yuanji Zhang\inst{1} \and Yuhao Huang\inst{1} \and Yuhan Zhang\inst{1} \and Xin Yang\inst{1} \and Ran Li\inst{2} \and Xuedong Deng\inst{3} \and Yanjun Liu\inst{4} \and Guowei Tao\inst{5} \and Yun Wu\inst{6} \and Sheng Zhao\inst{7} 
\and Xinru Gao\inst{8}$^{\href{mailto:gxr_ca@126.com}{\textrm{\Letter}}}$ 
\and Dong Ni\inst{1,9,10,11}$^{\href{mailto:nidong@szu.edu.cn}{\textrm{\Letter}}}$}  
\authorrunning{H. Liang et al.}
\institute{Medical Ultrasound Image Computing (MUSIC) Lab, School of Biomedical Engineering, Medical School, Shenzhen University, Shenzhen, China \and Shenzhen RayShape Medical Technology Co., Ltd, Shenzhen, China \and Center for Medical Ultrasound, The Affiliated Suzhou Hospital of Nanjing Medical University, Suzhou Municipal Hospital, Gusu School, Nanjing Medical University, Suzhou, China \and The First Hospital of China Medical University, Shenyang, China \and Qilu Hospital of Shandong University, Jinan, China \and Nanjing Women and Children's Healthcare Hospital, Women's Hospital of Naniing Medical University, Nanjing, China \and Maternal and Child Health Hospital of Hubei Province, Wuhan, China \and Northwest Women’s and Children’s Hospital, Xian, China \and School of Artificial Intelligence, Shenzhen University, Shenzhen, China \and National Engineering Laboratory for Big Data System Computing Technology, Shenzhen University, Shenzhen, China \and School of Biomedical Engineering and Informatics, Nanjing Medical University, Nanjing, China \\
    \email{nidong@szu.edu.cn; gxr\_ca@126.com}}

\maketitle              
\begin{abstract}

Fetal abdominal malformations are serious congenital anomalies that require accurate diagnosis to guide pregnancy management and reduce mortality. Although AI has demonstrated significant potential in medical diagnosis, its application to prenatal abdominal anomalies remains limited. Most existing studies focus on image-level classification and rely on standard plane localization, placing less emphasis on case-level diagnosis. 
In this paper, we develop a case-level multiple instance learning (MIL)-based method, free of standard plane localization, for classifying fetal abdominal anomalies in prenatal ultrasound. 
Our contribution is three-fold. 
First, we adopt a mixture-of-attention-experts module (MoAE) to weight different attention heads for various planes.
Secondly, we propose a medical-knowledge-driven feature selection module (MFS) to align image features with medical knowledge, performing self-supervised image token selection at the case-level. 
Finally, we propose a prompt-based prototype learning (PPL) to enhance the MFS. 
Extensively validated on a large prenatal abdominal ultrasound dataset containing 2,419 cases, with a total of 24,748 images and 6 categories, our proposed method outperforms the state-of-the-art competitors. Codes are available at: \url{https://github.com/LL-AC/AAcls}.

\keywords{Prenatal Abdominal Ultrasound  \and Multi-Instance Learning \and Anomalies Classification}

\end{abstract}
\section{Introduction}

Fetal abdominal malformations encompass a range of significant congenital anomalies that affect fetal health and are often associated with syndromes or aneuploidy~\cite{P2022}. The most severe and common abdominal anomalies include abdominal wall defects (such as omphalocele and gastroschisis), renal agenesis (RA), duodenal atresia (DA), and multicystic dysplastic kidney (MCDK), with an incidence of 1-4 per 10,000 live births~\cite{X2025,S2014,L2017}. Accurate diagnosis is critical for pregnancy decisions and treatment plans to reduce fetal morbidity and mortality. Ultrasound (US) remains the standard tool for prenatal abdominal evaluation, used to acquire standard planes and assess anatomical structures for diagnosing abdominal anomalies.
However, abdominal anomalies are diverse in type and presentation, often occurring at earlier gestational ages. The detection of these anomalies heavily relies on the expertise and experience of sonographers, leading to varying detection rates and underscoring the ongoing challenges in prenatal diagnosis.
To the best of our knowledge, artificial intelligence has demonstrated strong proficiency in diagnosing complex medical conditions~\cite{wang2020auto,huang2023fourier,liu2024mitral}. 
However, research on its application to the diagnosis of prenatal abdominal anomalies remains limited.

\textbf{Fetal Anomalies Detection}.
Currently, most studies for fetal anomalies detection primarily involve: 
(1) localizing standard planes from unordered \textbf{\textit{case-level}} ultrasound images or video sequences and
(2) detecting anomalies on the located \textbf{\textit{image-level}} standard planes. 
Xie et al.~\cite{Xie2020} classified the fetal brain standard planes, and then the lesions in anomaly images can be localized by activation map. 
Lin et al.~\cite{Lin2022} developed an intelligent system, using the YOLO3 to detect fetal brain standard planes and brain structure to classify abnormal images.
Qi et al.~\cite{Qi2025} constructed a deep learning model for nervous system anomalies classification in a multi-center fetal brain dataset.
Most recently, Ciobanu et al.~\cite{ciobanu2025automatic} proposed the first study to categorize fetal abdominal planes, achieving precise subsequent biomarker measurement and anomaly detection.

\begin{figure}[!]
\centering
\includegraphics[width=7cm]{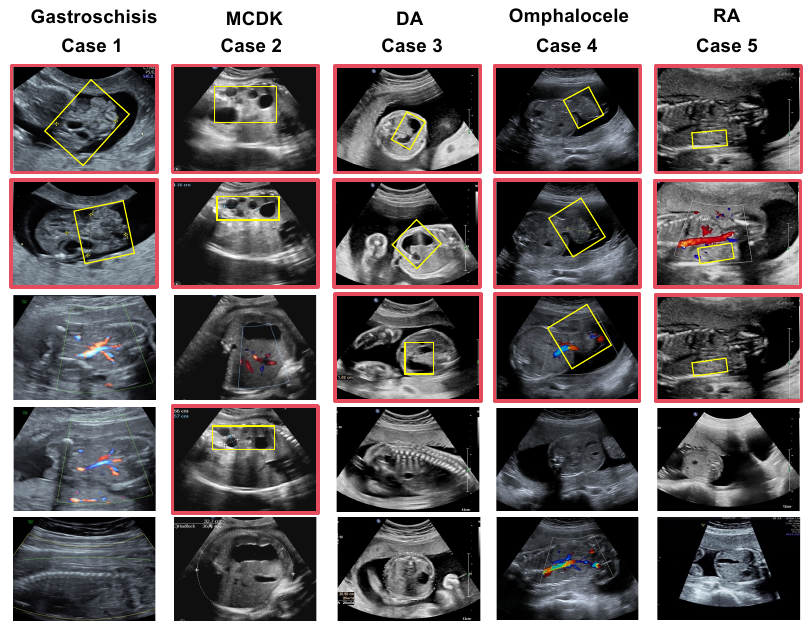}
\caption{\textbf{Dataset of Multi-Center Prenatal Abdominal Ultrasound}. Each case contains several ultrasound images taken during a single examination, including both abnormal and normal images. Abnormal images are annotated with red boxes. Abnormal structures are annotated with yellow boxes.}
\label{fig1}
\end{figure}

Although effective, the above studies require standard plane localization, which necessitates fine-grained image-level annotations. Meanwhile, they face challenges in cases such as abdominal wall defects (e.g., gastroschisis and omphalocele, shown in Fig.~\ref{fig1}), where organ displacement or external exposure leads to the loss of abdominal structures, resulting in the failure to locate standard planes.
To address these issues, we propose a Multiple Instance Learning (MIL)-based method that only requires case-level annotations (coarse-grained annotations from an ultrasound image pool taken from a single examination) and does not rely on localizing standard planes. 

\textbf{Multiple Instance Learning (MIL)} represents a powerful tool for weakly supervised classification tasks. 
Specifically, it aims to aggregate multiple instances (image-level) into bag-level labels (case-level)~\cite{Additive,DTMIL,MAX,TransMIL,DSMIL,IBMIL,Mamba2,Mamba,ABMIL}. 
Zhuang et al.~\cite{Zhuang2024} proposed a multi-scale feature fusion-based attention MIL model for case-level classification of thyroid nodules. 
Kaito et al.~\cite{SAMIL} proposed a selective aggregated MIL for case-level severity in ulcerative colitis diagnosis. Wang et al.~\cite{Wang2023} proposed a deep semi-supervised MIL framework for classifying diabetic macular edema from optical coherence tomography images.

The goal of this paper is to develop an MIL-based method that does not rely on localizing standard planes for case-level prenatal abdominal ultrasound anomaly classification. Our contribution is three-fold.
(1) We adopt a mixture-of-attention-experts module (MoAE): By adopting MoAE, we can assign specific attention experts for different planes (i.e., sagittal, coronal, and transverse planes).
(2) We propose a medical-knowledge-driven feature selection module: A self-supervised feature selection mechanism is designed based on medical knowledge, reducing the dependency on standard planes.
(3) We propose a prompt-based prototype learning method to differentiate features across different categories.
Experiments comparing our method with SOTA MIL methods show significant improvements in abdominal anomaly classification, providing a reliable and automated solution for prenatal ultrasound screening.

\section{Methodology}
\begin{figure}[!tp]
\includegraphics[width=\textwidth]{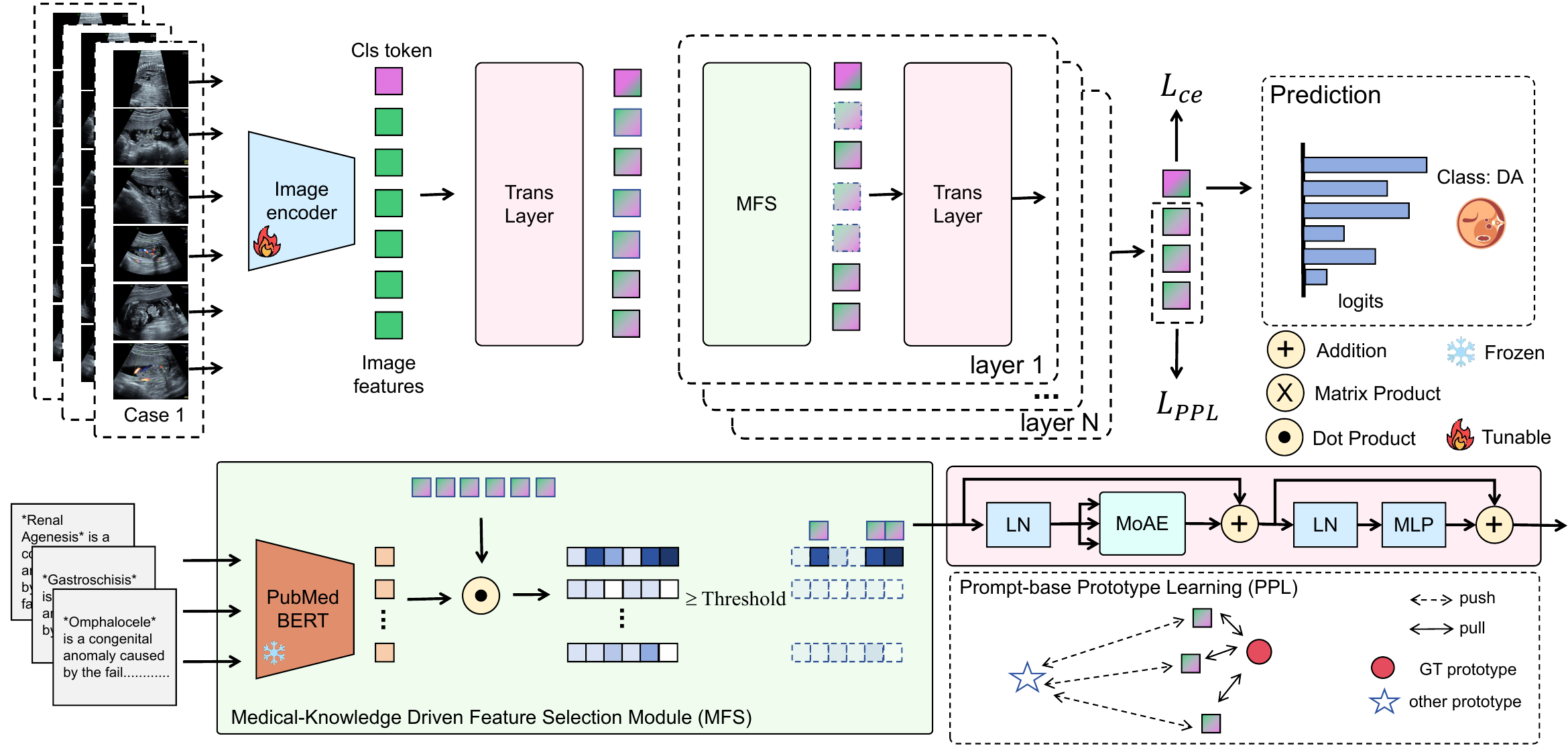}
\caption{Overview of our proposed framework}
\label{fig2}
\end{figure}

Fig.~\ref{fig2} shows an overview of our proposed framework. First, a case image pool is fed into transformer layers for image integration. We adopt a mixture-of-attention-experts module (MoAE), which replaces the vanilla multi-head attention mechanism to control different attention experts, enabling feature integration across different planes. Secondly, a medical-knowledge-driven feature selection module (MFS) leverages medical knowledge to select significant images from the case image pool. Finally, the prompt-based prototype learning (PPL) module involves a loss function to bring paired medical knowledge and images closer in the feature space, ensuring the distinctiveness of the selected image.

\textbf{TransMIL Definition for Case-level Analysis.} The vanilla TransMIL~\cite{TransMIL} was designed for Whole Slide Imaging (WSI), including two main components: the image encoder and the transformer layers. 
Due to the large size, WSIs are usually divided into multiple non-overlapping patches as instances. 
The image encoder $f$ first extracts features $X'$ from these instances $X$, which are then fed into the transformer layers to generate the final prediction $y$. 
In our task, we treat a image pool $X$ obtained from a single examination of a case as instances, each instance $x_i$ represents one image, formulated by:
\begin{equation}
X' = f(X), X = ({x_1, x_2, \dots, x_n}),
\end{equation}
\begin{equation}
y = TransLayer(h_{cls}, X'), X' = (x'_1, x'_2, \dots, x'_n).
\end{equation}

\begin{figure}[!tp]
\centering
\includegraphics[width=0.45\textwidth]{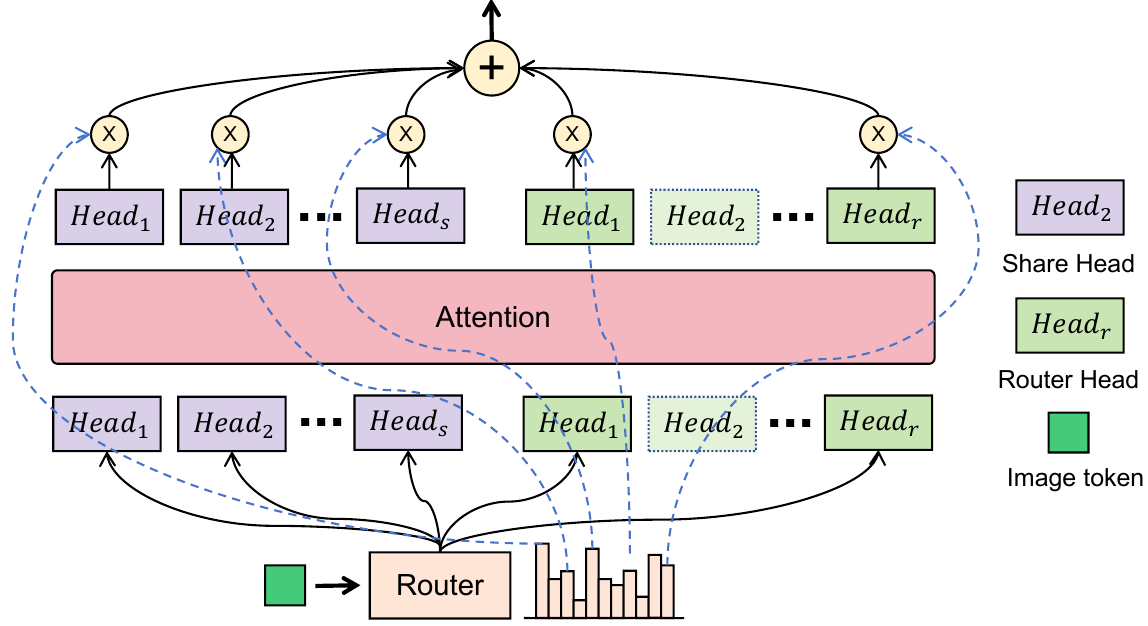}
\caption{Mixture of Attention Experts Module}
\label{fig2-1}
\end{figure}

\textbf{Mixture of Attention Experts Module for Integrating Planes.} 
For the transformer layer, multi-head attention is crucial, and different attention heads have different functions, including positional function~\cite{V2019}.
Here, we consider different planes to possess spatial position information.
We adopt the mixture-of-attention-experts module (MoAE)~\cite{MOH} as shown in Fig.~\ref{fig2-1}, treating attention heads as experts within the MoE mechanism, enabling feature integration across different planes, formulated by:

\begin{equation}
    MoAE(X') = \sum^h_{i=1}g_iH^iW^i_O,
\end{equation}
\begin{equation}
    H^i = Attention(X'W^Q, X'W^k, X'W^v),
\end{equation}
\begin{equation}
    \begin{aligned}
        g_i =
        \begin{cases}
        \alpha_1 \text{Softmax}(W_s x'_i)_j, & \text{if } 1 \leq j \leq h_s, \\
        \alpha_2 \text{Softmax}(W_r x'_i)_j, & \text{if } (W_r x'_i)_j \in \text{Top-K}(\{(W_r x'_i)_j | h_s + 1 \leq j \leq h\}), \\
        0, & \text{otherwise.}
        \end{cases}
    \end{aligned} 
\end{equation}
where \( H^i \) represents the output of the vanilla Attention mechanism. $g_i$ represents the router score.  Each token activates the top-K most relevant attention heads via routers. When the $i_{th}$ attention head is activated, $g_i > 0$. $ Softmax(W_rx'_i) $ represents the router score. $W$ represents a trainable projection matrix. This module also retains shared heads $h_s$ to integrate common information across tokens. And the coefficients $ \alpha_1$ and $\alpha_2$ balance the contributions of the shared and routed heads. This method enables each token to select the most relevant attention heads and produce a weighted summation based on router scores.

\textbf{Medical-Knowledge Driven Feature Selection Module.} Recent multi-modal pre-training models, such as CLIP~\cite{CLIP}, MedCLIP~\cite{MedClip} and BioMedCLIP~\cite{BioMedClip}, have learned associations between images-text by training on large-scale language-image pairs using contrastive learning, aligning them in the feature space and enabling zero-shot or few-shot learning capabilities. Additionally, prompt learning has utilized a pre-trained Vision-Language Model (VLM) to adapt to new downstream tasks by incorporating task-specific prompts into the model's input, demonstrating strong performance~\cite{C2023,Y2023}. 

Inspired by these, we propose a medical-knowledge-driven feature selection module (MFS).
Guided by medical-knowledge prompts, The MFS performs self-supervised selection of abnormal images from the case-level image pool without relying on standard plane localization.

First, medical-knowledge are fed into a pre-trained model PubMedBERT~\cite{PubMedBERT} to extract the sentence feature as the $prompt$. The input format of the $prompts$ is: $ Prompt = \{Cls, definition, signs\} $. 
Secondly, we compute the similarity between each image token and the medical-knowledge $prompts$ to obtain an activation score. Then, we use an adaptive threshold to filter out irrelevant images.
The $\beta$ increases progressively at each stage, the formula of \textit{Threshold} can be written by:

\begin{equation}
    \begin{aligned}
        Threshold = \beta*\frac{1}{n}\sum^n_{i=1}max(Softmax(x'_i \cdot Prompts), dim=C).
    \end{aligned}
\end{equation}

\textbf{Prompt-base Prototype learning for Enhancing MFS.} To ensure that the images selected by MFS are representative, we propose a loss function based on the image tokens $X'$ as follows:
\begin{equation}
    \begin{aligned}
        L_{PPL}(X', Prompts, Cls) =  \sum^{N}(\frac{1}{n}\sum^n_{j=1}\frac{exp(x'_j \cdot Prompts_{Cls})}{\sum^C_{i=1} exp(x'_j \cdot Prompts_i)})
    \end{aligned},
\end{equation}
where $Prompts$ represents the set of all category prompts, $ C=len(Prompts) $ and $Cls$ denotes the ground truth. $N$ represents the size of the dataset, while $n$ represents the number of images in a single case. The image token $x'_j$ selected by MFS and the category prompt are used as a positive pair, while other prompts form negative pairs.
Treating prompts as prototypes, we pull the image tokens of the positive pair closer to the prototype center and push them away from other centers, further enhancing the effectiveness of the selected features in MFS.
We further add cross-entropy loss, and the overall loss function is:
\begin{equation}
    L(y, Cls, Prompt) = L_{ce}(y, Cls) + L_{PPL}(X', Prompts, Cls).
\end{equation}

\section{Experimental Results}

\textbf{Ethical Approval.} This study was approved by the Clinical Research Ethics Committee of the Affiliated Suzhou Hospital of Nanjing Medical University (K-2023-067-H01) and conducted following the Declaration of Helsinki. It was registered at \url{www.chictr.org.cn} (ChiCTR2300071300).

\textbf{Materials and Implementation Details.} In this paper, we collected a multi-center prenatal abdominal ultrasound dataset.
The age range of the pregnant women was 17 to 48 years, and the gestational age ranged from 10 to 39 weeks. 
Our dataset contains 2,419 cases and a total of 24,748 images, including abdominal wall defect and several common urological, digestive system anomalies: Duodenal Atresia (DA) (476), Gastroschisis (154), Omphalocele (760), Renal Agenesis (RA) (197), Multicystic Dysplastic Kidney (MCDK) (277), and Normal (555).
Each case contains 1 to 65 images.
The dataset was randomly split into training (1,516), validation (449), and test (454) sets at the case-level, ensuring that the distribution of categories was maintained across all subsets.

We implemented our method in PyTorch, using an NVIDIA 4090 GPU. The batch size, learning rate, and total epochs were set to 64, 1e-3, and 100, respectively. We used the Adam optimizer for model optimization. We used pre-trained model ResNet18 as image encoder. During training, we employed a warm-up learning rate strategy, starting with an initial learning rate of 1e-8, reaching the maximum learning rate at epoch 20, and then gradually decreasing using a half-cosine schedule to a minimum of 1e-8.

\begin{table}[!ht]
    \centering
    \caption{Method comparison. The best results are shown in bold.} 
    \label{table1}
    \renewcommand{\arraystretch}{1.5}
    \scalebox{0.65}{
    \begin{tabular}{cccccccccccccc}
    \hline
        ~ & \multicolumn{2}{c}{DA} & \multicolumn{2}{c}{Gastroschisis} & \multicolumn{2}{c}{Omphalocele} & \multicolumn{2}{c}{RA} & \multicolumn{2}{c}{MCDK} & \multicolumn{2}{c}{Normal} & \multirow{2}*{Acc} \\ 
        ~ & Sen & F1 & Sen & F1 & Sen & F1 & Sen & F1 & Sen & F1 & Sen & F1 & ~ \\ 
        \hline
        Features Max & 87.50\% & 91.30\% & 46.88\% & 50.00\% & 82.89\% & 84.56\% & 75.00\% & 74.07\% & 82.14\% & 79.31\% & \textbf{97.44\%} & 89.94\% & 78.64\% \\ 
        Features Mean & 92.71\% & 92.71\% & 43.75\% & 46.67\% & 78.95\% & 81.91\% & 70.00\% & 74.67\% & \textbf{87.50\%} & 82.35\% & 96.15\% & 88.76\% & 78.18\% \\ 
        TransMIL & 93.75\% & 94.74\% & 53.12\% & 65.38\% & \textbf{93.42\%} & 86.59\% & 67.50\% & 72.00\% & 73.21\% & 82.83\% & 96.15\% & 91.46\% & 79.53\% \\ 
        AbMIL & 91.67\% & 89.34\% & 50.00\% & 57.14\% & 87.50\% & \textbf{87.79\%} & 77.50\% & \textbf{79.49\%} & 85.71\% & 88.89\% & 94.87\% & 89.16\% & 81.21\% \\ 
        DSMIL & \textbf{97.92\%} & 95.92\% & 46.88\% & 58.82\% & 90.79\% & 87.34\% & 72.50\% & 79.45\% & 85.71\% & 88.89\% & 93.59\% & 89.02\% & 81.23\% \\ 
        AdditiveMIL & 90.62\% & 92.06\% & 50.00\% & 54.24\% & 83.55\% & 85.52\% & 67.50\% & 69.23\% & \textbf{87.50\%} & 83.76\% & 94.87\% & 88.10\% & 79.01\% \\
        IBMIL & 93.75\% & 92.78\% & 43.75\% & 53.85\% & 86.18\% & 85.62\% & 80.00\% & 72.73\% & 76.79\% & 81.90\% & 92.31\% & 88.34\% & 78.80\% \\  
        Mamba2MIL & 94.79\% & 96.3\% & 53.12\% & 56.67\% & 78.95\% & 83.33\% & 77.50\% & 61.39\% & 73.21\% & 81.19\% & 93.59\% & 86.39\% & 78.53\% \\
        SAMIL & 92.71\% & 89.90\% & 40.62\% & 44.83\% & 79.61\% & 81.48\% & 80.00\% & 79.01\% & 80.36\% & 79.65\% & 87.18\% & 84.47\% & 76.75\% \\ 
        Our & \textbf{97.92\%} & \textbf{97.92\%} & \textbf{62.50\%} & \textbf{67.80\%} & 87.50\% & \textbf{87.79\%} & \textbf{85.00\%} & 73.91\% & 80.36\% & \textbf{89.11\%} & \textbf{97.44\%} & \textbf{94.41\%} & \textbf{85.12\%} \\ 
        \hline
        \end{tabular}
        }
\end{table}

\begin{table}[!ht]
    \centering
    \caption{Ablation studies. The best results are shown in bold.} 
    \label{table2}
    \renewcommand{\arraystretch}{1.5}
    \scalebox{0.65}{
    \begin{tabular}{ccccccccccccccccc}
    \hline
        \multirow{2}*{MFS} & \multirow{2}*{MoAE} & \multirow{2}*{PPL} & \multicolumn{2}{c}{DA} & \multicolumn{2}{c}{Gastroschisis} & \multicolumn{2}{c}{Omphalocele} & \multicolumn{2}{c}{RA} & \multicolumn{2}{c}{MCDK} & \multicolumn{2}{c}{Normal} & \multirow{2}*{Acc} \\ 
        
        ~ & ~ & ~ & Sen & F1 & Sen & F1 & Sen & F1 & Sen & F1 & Sen & F1 & Sen & F1 & ~ \\
        \hline
        \XSolidBrush & \XSolidBrush & \XSolidBrush & 93.75\% & 94.74\% & 53.12\% & 65.38\% & \textbf{93.42\%} & 86.59\% & 67.50\% & 72.00\% & 73.21\% & 82.83\% & 96.15\% & 91.46\% & 79.53\% \\ 
        \Checkmark & \XSolidBrush & \XSolidBrush & 96.88\% & 97.38\% & 56.25\% & 58.06\% & 82.89\% & 84.85\% & 72.50\% & 69.05\% & \textbf{85.71\%} & 88.89\% & 96.15\% & 90.36\% & 81.73\% \\ 
        \Checkmark & \XSolidBrush & \Checkmark & 95.83\% & 96.34\% & \textbf{62.50\%} & 59.70\% & 85.53\% & 86.09\% & 75.00\% & \textbf{78.95\%} & 83.93\% & 87.85\% & \textbf{1.00\%} & \textbf{94.55\%} & 83.80\% \\ 
        \Checkmark & \Checkmark & \Checkmark & \textbf{97.92\%} & \textbf{97.92\%} & \textbf{62.50\%} & \textbf{67.80\%} & 87.50\% & \textbf{87.79\%} & \textbf{85.00\%} & 73.91\% & 80.36\% & \textbf{89.11\%} & 97.44\% & 94.41\% & \textbf{85.12\%} \\ \hline
    \end{tabular}
    }
\end{table}

\textbf{Result.} Table \ref{table1} presents a comparison of our method with nine strong competitors, including Feature Max \cite{MAX}, Feature Mean \cite{MAX}, TransMIL \cite{TransMIL}, AbMIL \cite{ABMIL}, DSMIL \cite{DSMIL}, AdditiveMIL \cite{Additive}, IBMIL \cite{IBMIL}, Mamba2MIL \cite{Mamba2}, and SAMIL \cite{SAMIL}. 
In anomalies classification, we primarily focus on the sensitivity for detecting abnormal fetuses. Therefore, we show sensitivity as the performance metric for each category, while also presenting the F1 score. Further, we use weighted accuracy as the overall performance metric to account for class imbalance in the dataset.
Results show that our method outperforms all other methods in most categories. 
Regarding sensitivity improvement, compared to the second-best model, the improvements are DA (3.13\%), Gastroschisis (9.38\%), RA (5\%), and Normal (1.29\%), respectively.
In terms of F1 score, the improvements are DA (1.62\%), Gastroschisis (2.42\%), Omphalocele (0.45\%), MCDK (0.22\%), and Normal (2.95\%), respectively.
Additionally, our method achieves the best overall accuracy 85.12\% among all the methods, demonstrating its effectiveness.

\begin{figure}[!ht]
\centering
\includegraphics[width=\textwidth]{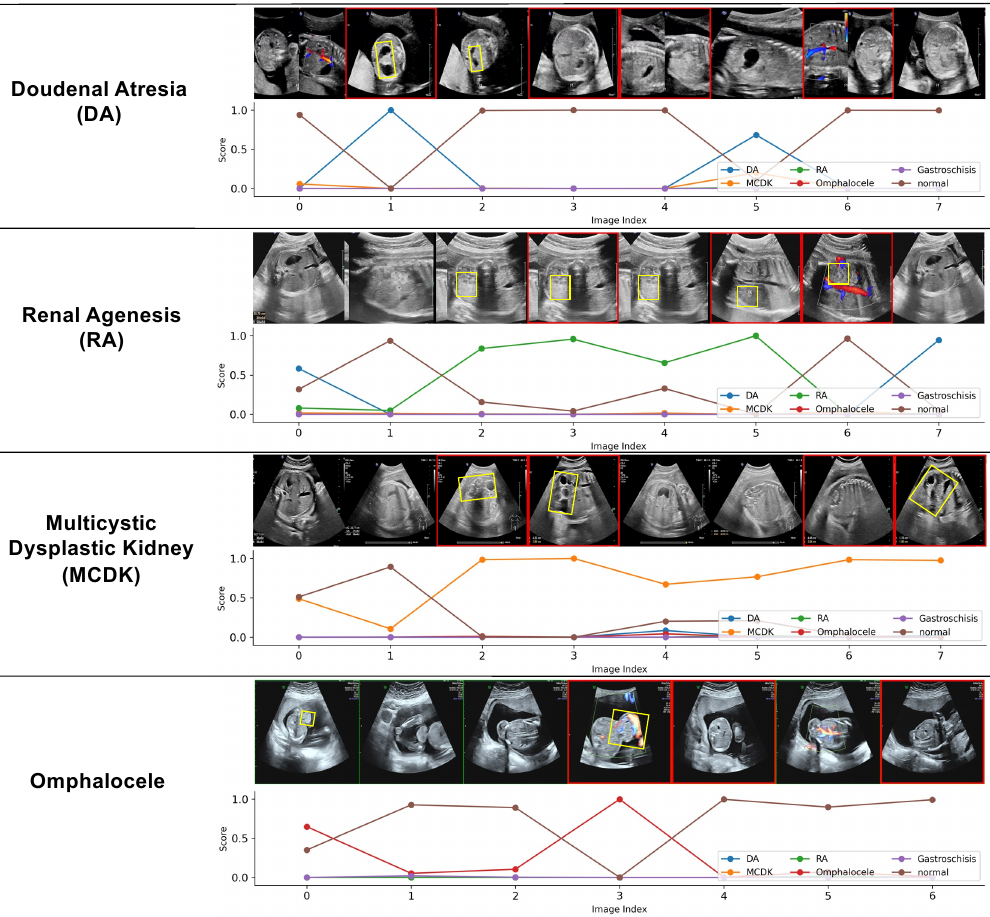}
\caption{\textbf{Visualization of typical cases.} Abnormal structures are annotated with yellow boxes. The images activated by MFS are annotated with red boxes. The line chart represents the activation scores of the corresponding image categories.}
\label{fig3}
\end{figure}

We also preformed ablation studies to evaluate the contribution of each proposed module in Table \ref{table2}. Specifically, we treated the TransMIL as the baseline and added our proposed modules. 
When using the MFS module alone, it can be seen that MFS contributes about 2\% improvement to the overall accuracy. Furthermore, the combination of MFS and PPL further enhances performance, resulting in an additional 2\% improvement. As the modules were progressively added, the overall accuracy increased significant of 5.59\%, highlighting the effectiveness of the proposed modules in enhancing model performance.

Fig.~\ref{fig3} displays typical cases using our modules. The images annotated with red boxes indicate those retained after applying our MFS, while the rest are discarded. As the number of image tokens decreases, the model becomes more focused on anomaly classification. The line chart shows that the images activated by MFS generally receive higher scores. This demonstrates that even if the model does not select all abnormal images, the chosen ones remain highly representative as shown in Fig.~\ref{fig3} (RA and Omphalocele). These typical cases demonstrate that our method can perform case-level predictions effectively without relying on standard plane localization.

\section{Conclusion}
In this paper, we propose a medical-knowledge driven MIL method for improving case-level prenatal abdominal ultrasound anomalies classification. 
We first adopt MoAE, assigning different attention experts to specific planes enabling feature integration. 
Next, we propose the MFS, which enables the model to perform self-supervised selection of significant images from case-level ultrasound image pool based on medical knowledge.
Finally, we propose a loss function PPL to enhance the MFS. Extensive experiments show the effectiveness of our method. In the future, we will extend this framework to more congenital anomalies classification in fetus.

\begin{credits}
\subsubsection{\ackname} This work was supported by the grant from National Natural Science Foundation of China (12326619, 62171290); Science and Technology Planning Project of Guangdong Province (2023A0505020002); Frontier Technology Development Program of Jiangsu Province (BF2024078); Science and Technology Research and Development Program of Shaanxi Province (2011K12-05-11); Key Research and Development Program of Shaanxi (2025SF-YBXM-156).

\subsubsection{\discintname} The authors have no competing interests to declare that are relevant to the content of this article.

\end{credits}

%
%

\bibliographystyle{splncs04}
\bibliography{reference}

\end{document}